\documentclass[letterpaper,12pt]{article}

\usepackage{times}
\usepackage{booktabs} % For formal tables
\usepackage{amsfonts,color,amssymb,amsmath}
\usepackage{graphicx,algorithmic,algorithm2e}

\begin{document}

\pagestyle{empty}
\begin{tabular}{l}
{\bf A Grid Based Adversarial Clustering Algorithm} \\[12pt] 
Wutao~Wei,~Nikhil~Gupta,~Bowei~Xi, \\[12pt]
Purdue University, \\[12pt]
Corresponding to: \\
Bowei Xi, Department of Statistics, Purdue
University, \\
West Lafayette, IN, 47907. xbw@purdue.edu \\[38pt]
\end{tabular}

\noindent
{\bf Abstract}: Nowadays more and more data are gathered for
detecting and preventing cyber attacks. In 
cyber security applications, data 
analytics techniques have to deal with active adversaries that
try to deceive the data analytics models and avoid being
detected. The existence of
such adversarial behavior motivates the development of 
robust and resilient adversarial learning techniques for various 
tasks.  Most of the previous work focused on adversarial
classification techniques, which assumed the
existence of a reasonably large amount of carefully labeled data
instances. However, in practice, labeling the data instances
often requires costly 
and time-consuming human expertise and becomes a significant
bottleneck. Meanwhile, a large number of unlabeled instances
can also be used to understand the adversaries' behavior.
To address the above mentioned challenges,
in this paper, we develop a novel 
grid based adversarial clustering algorithm. Our adversarial
clustering algorithm is able to identify the core normal regions,
and to draw defensive walls around the centers of the
normal objects utilizing game theoretic ideas. Our algorithm also
identifies sub-clusters of attack objects, the overlapping areas within  
clusters, and outliers which may be potential anomalies.   


\section{Introduction} 
\label{sec:intro}
Increasingly data analytics techniques are being applied to large volumes of
system monitoring data to detect cyber security incidents. 
The ultimate goal is to provide cyber
security analysts with robust and effective insights derived
from big data.  
Unlike most other application domains, 
cyber security applications often face adversaries who
actively modify their strategies to launch new and unexpected
attacks. The existence of such adversaries results in 
cyber security data that have  unique
properties. Firstly, the attack instances are frequently being
modified to avoid detection. Hence a future dataset no longer
shares the same properties as the current training dataset. 
Secondly, when a previously unknown attack happens, security
analysts need to respond to the new attack quickly without the
help of readily labeled instances in their database to assist their work.
Thirdly, adversaries can be well funded and make big investments to
camouflage the attack instances. Therefore despite significant effort invested by the domain experts, a vast majority of the instances in
their database may remain
unlabeled. For example, a new malware can incorporate large amounts of
legitimate code to masquerade as legitimate software and
obfuscate its binary.  In other cases, it may
become laborious and expensive to label an instance.   

Thus data analytics techniques for cyber security must also have
unique capabilities. They 
need to be resilient against
the adaptive behavior of the adversaries, and are able to quickly
detect previously unknown and unlabeled new attack instances. 
Hence, recently, various 
adversarial machine learning techniques 
have been developed to counter adversarial adaptive behaviors. 
However those previous adversarial
machine learning work is mostly under the main assumption of 
the availability of large amounts of labeled instances (i.e., normal
versus malicious objects). Although large amounts of data are  
generated by the cyber security applications, we often have few
properly labeled instances to construct an effective classifier.  
                      
Given a large
amount of unlabeled data, defender needs to capture the
adversarial behavior, identify suspicious instances as anomalies for a more
detailed investigation, and quickly respond to new attacks. 
However clusters identified by traditional clustering algorithms are
likely mixed, since with a few attack objects, adversaries can
bridge the gap between two previously well separated
clusters. 
Sometimes a handful of labeled attack and
normal instances are available. There are too few of them to build a
classifier, yet they offer valuable information about the
adversaries.  In this paper, we develop a novel
adversarial clustering algorithm so that we need only a few
labeled instances to build robust defensive algorithm against the
attack objects. Our algorithm can identify the centers of normal
objects, sub-clusters of attack objects, and the overlapping
areas where adversaries have successfully
placed the attack objects. We then draw defensive walls
around the centers of the normal objects utilizing game
theoretic ideas. Our algorithm also identifies
outliers as potential anomalies and outlying 
unknown clusters for further investigation.   

Semi-supervised learning techniques also utilize information from
both labeled and unlabeled instances. Adversarial
clustering and semi-supervised learning 
operate under very different assumptions. In adversarial
settings, attackers purposely modify the attack objects to make
them similar to normal objects, though suffering a cost for doing
so. Hence the assumptions commonly used for semi supervised
learning do not hold for adversarial clustering. Instead we
observe that objects
similar to each other belong to different classes, while objects
in different clusters belong to the same class. In adversarial
settings, within each cluster, objects from two classes can overlap
significantly. Consequently, adversarial clustering and
semi-supervised learning techniques have 
very different goals too. Semi-supervised learning aims to assign
labels to all the unlabeled objects with the best accuracy. Our adversarial
clustering algorithm aims to identify the overlapping regions, and 
the core areas of the normal objects, within each 
cluster. The overlapping regions and outliers are not labeled by our
algorithm. 
%MK: Next sentence is not clear. Please consider re-writing
We draw defensive walls around the centers of the normal objects. 
The shape and the size of the defensive walls are determined
through a game theoretic study. 
Inside the defensive walls, we have nearly pure
normal objects, despite an increased error of blocking out the normal
objects mixed with the attack objects 
outside of the walls. Adversarial clustering draws an analogy to
airport security. A small number of passengers use the fast
pre-check lane at the security checkpoint, analogous to the
normal objects inside the defensive walls. All other passengers
must go through more time consuming security check, analogous to
the objects outside the walls. The goal is not to let a single
terrorist enter an airport, at a cost of blocking out many normal 
objects. Meanwhile the ability to identify the overlapping regions leads to a
more focused security check procedure, where attack and normal 
objects are similar to each other. We compare our algorithm with
semi-supervised learning algorithms in Section~\ref{sec:exp}. 

The paper is organized as follows. Section~\ref{sec:related}
discusses the related work. In Section~\ref{sec:advl-cluster},
we present our adversarial clustering algorithm. In
Section~\ref{sec:game}, we conduct a game theoretic study to
examine the sizes and the shapes of different defensive walls
used in our adversarial clustering algorithm. In
Section~\ref{sec:exp}, we evaluate our algorithm
with simulated and a network intrusion
data sets. Section~\ref{sec:conclusion} concludes the paper.  

\subsection{Related Work}
\label{sec:related}

Robust learning techniques have been proposed in the past, 
for example, to defeat poisoning attacks \cite{poison-2009}, purposely
generated malicious errors \cite{kearns1993}, and missing or
corrupted features \cite{corrupted-class-2010}.  
Classification in adversarial settings has also received considerable
attention in the literature, e.g.,  
\cite{Domingos-2004-advl-classification,advlNelson-2011,phishing-advl-2009,lowd-2005,xi-2013}.  
In \cite{Scheffer-2011,xi-2011}, 
Stackelberg game is used to model the sequential actions between
a defender/classifier and active adversaries.  
Adversarial classification techniques were developed for the
Facebook social network to defeat the fake and spam accounts 
\cite{facebook2011}. 
However even with the information obtained
from a large training sample of labeled normal and attack
objects, building a robust classifier to block out the attack
objects, which are constantly being modified by adversaries to avoid
detection, is not an easy task. 
 
Compared with adversarial classification, there is fewer work on
adversarial clustering, which is a much harder learning problem.  
\cite{biggio-cluster-secure-2013} considered the problem of
evaluating the security of clustering algorithms in an adversarial
setting. \cite{biggio-cluster-secure-2013} then 
evaluated the security of single linkage hierarchical clustering
algorithm under  poisoning attacks and obfuscation attacks. 
\cite{biggio-2014} further studied the effects of poisoning
attacks on complete linkage hierarchical clustering
algorithm. \cite{hide2008} showed that a few well-constructed
attack objects could lead to a larger mixed cluster, and hence
significantly reduce the effectiveness of a 
clustering algorithm. \cite{icml2013} showed that subspace
clustering has a certain tolerance for noisy or corrupted data. 

Semi-supervised learning techniques utilizes information from
both labeled and unlabeled instances. It has an extensive
literature. In general
there are two types of semi-supervised learning
techniques, semi-supervised classification and semi-supervised
clustering. There are many different approaches for
semi-supervised classification, such as transductive support vector
machine (TSVM), generative mixture
models, self-training and co-training. 
TSVM extends SVM to the semi-supervised learning scenario. Labels
are assigned to the unlabeled instances such that the
classification boundary has the maximum margin on the original
labels and newly assigned labels
(e.g.,~\cite{semiclass-tsvm1,semiclass-tsvm2,semiclass-tsvm3}). TSVM
avoids the high density regions, which may not be the optimal
solution when two classes are heavily overlapped. 
Under mixture model
assumption, EM algorithm is used for semi-supervised
classification (e.g.,~\cite{semiclass-mix1,semiclass-mix2}). This
approach allows the classification boundary to go through the
densest region of the data points. However users need to pay
attention to model identifiability issue and whether the model
assumption fits the data or not
(e.g.,~\cite{semiclass-mix3,semiclass-mix4,semiclass-mix5}).       
Self-training approach iteratively assigns labels to new data
points, and then includes both the existing labels and newly
assigned labels to train another classifier
(e.g.,~\cite{semiclass-self1,semiclass-self2,semiclass-self3}). 
Co-training splits the available features into two sets and build two
classifiers, each using only one set of features. In an iterative
process, each classifier learns from the other one with the most
confident predicted labels
(e.g.,~\cite{semiclass-co1,semiclass-co2,semiclass-co3}). 

Often semi-supervised
clustering algorithms use pairwise must-link and cannot-link
constraints. Must-links ensure the objects with identical labels 
are grouped in the same cluster, while cannot-links ensure the
objects with different labels are in different clusters
(e.g.~\cite{semiclu1,semiclu2,semiclu3}). Meanwhile many work extends
K-means algorithm to semi-supervised clustering settings
(e.g.,~\cite{semiclu4,semiclu5,semiclu6}). 
\cite{semihiss} developed a hierarchical density based
semi-supervised clustering algorithm. However if the density
varies significantly among clusters, the algorithm has difficulty
to extract the natural cluster structure. 
\cite{semidb} extends DBSCAN to semi-supervised settings. Instead
of having one set of values for the parameters as in DBSCAN,
\cite{semidb} finds multiple sets of parameter values to better handle the
situation when densities vary significantly among clusters. 

Our adversarial clustering algorithm has a very different
goal. Compared with semi-supervise learning, 
we do not label all the previously unlabeled objects and
attempt to achieve the maximum accuracy. Instead we identify the
centers of normal objects using defensive walls. We focus on having
nearly pure normal objects inside the walls, often at the expense
of blocking out many normal objects mixed with abnormal
objects. Hence the overall accuracy of our algorithm may decrease
but we identify the center normal regions where the percentage of
normal objects is much higher, and can be considered as
relatively safe regions. 
We do not label the objects in the regions where normal
and abnormal objects are mixed. Instead we mark out the whole
mixed areas, where attacks take place and objects must be
examined carefully. We also leave unknown clusters and outliers
unlabeled, since they should be investigated carefully as being
potential anomalies or a new attack. 

%MK: I think adding few sentences on why our work is different from all these work would be useful.

%%
\section{Adversarial Clustering} 
\label{sec:advl-cluster}

In cyber security applications, adversaries actively manipulate
the objects under their control to break through a defensive
algorithm. Hence the properties of the data under attack are
drastically different from the data without an attack. 
Even though the normal population remains
unchanged, the
adversaries can inject a small amount of attack objects to
fill in the gap between abnormal clusters and normal clusters, 
and make previously relatively pure normal clusters mixed, as
pointed out in
\cite{biggio-cluster-secure-2013,biggio-2014,hide2008}.  
Traditional clustering algorithms are able to produce clusters
and a few outliers. Without any labeled instances, that is the
only result we can expect, not knowing whether a cluster is
mixed, or nearly purely normal or abnormal. On the other hand, if a large number of
labeled instances are available, we can build a classifier with a
well defined classification boundary that separates the normal and
abnormal objects within one cluster, and separate the relative pure
normal clusters from the abnormal objects.  

In this paper, we consider a scenario where there are a large
number of unlabeled instances and only a handful of labeled
instances (i.e., the number labeled being far less than the number of
unlabeled ones). A classifier created using very few labeled
objects is very inaccurate when being applied to the large number
of unlabeled ones. 
On the other hand clusters, produced by
traditional clustering algorithms, may become mixed clusters
under attack, where extra efforts are needed to identify normal
and abnormal regions inside these mixed clusters.   
Therefore we develop a grid based
adversarial clustering algorithm, which is able to utilize the
handful of labeled objects, identify relatively pure
normal and abnormal regions within one cluster and their overlapping area,
and further identify outliers and outlying clusters which
need more effort to investigate their properties.  

A classifier with a well defined classification boundary is
analogous to a point estimate. When the sample size is too small
(i.e., too few labeled instances), a point estimate is way too
inaccurate. Hence our clustering algorithm identify 
overlapping areas between the normal regions and the abnormal regions,
analogous to confidence regions. 
When a large number of labeled instances
are available, a classification boundary is a
defensive wall against the adversaries, since it blocks out the
attack objects. When facing active adversaries, a classifier needs to be more
conservative, i.e., a classification boundary is pulled back
toward the center of the normal population, as shown
in~\cite{xi-2013}. With a large number of unlabeled instances, our
adversarial clustering algorithm offers more valuable information
to capture both normal and abnormal regions. Our adversarial
clustering algorithm then 
plays a conservative strategy. We draw defensive walls inside 
the normal regions to protect the relatively pure 
normal centers.  
All objects outside of the
walls need to be examined carefully, while many normal objects
can be blocked out. How conservative the defensive walls need to
be is determined through a game theoretic study in
Section~\ref{sec:game}. If the defensive walls are too close to
the center of the normal regions, we miss a large portion of the
relatively pure normal areas. If the defensive walls are too
relaxed, we have too many attack objects in the walls. Hence we 
utilize the equilibrium
information to determine the sizes of the conservative defensive
walls for our algorithm.     
 
\subsection{A Grid Based Defensive Clustering Algorithm}

\begin{figure*}[tb!]
\centering{
\includegraphics[width=1.3in]{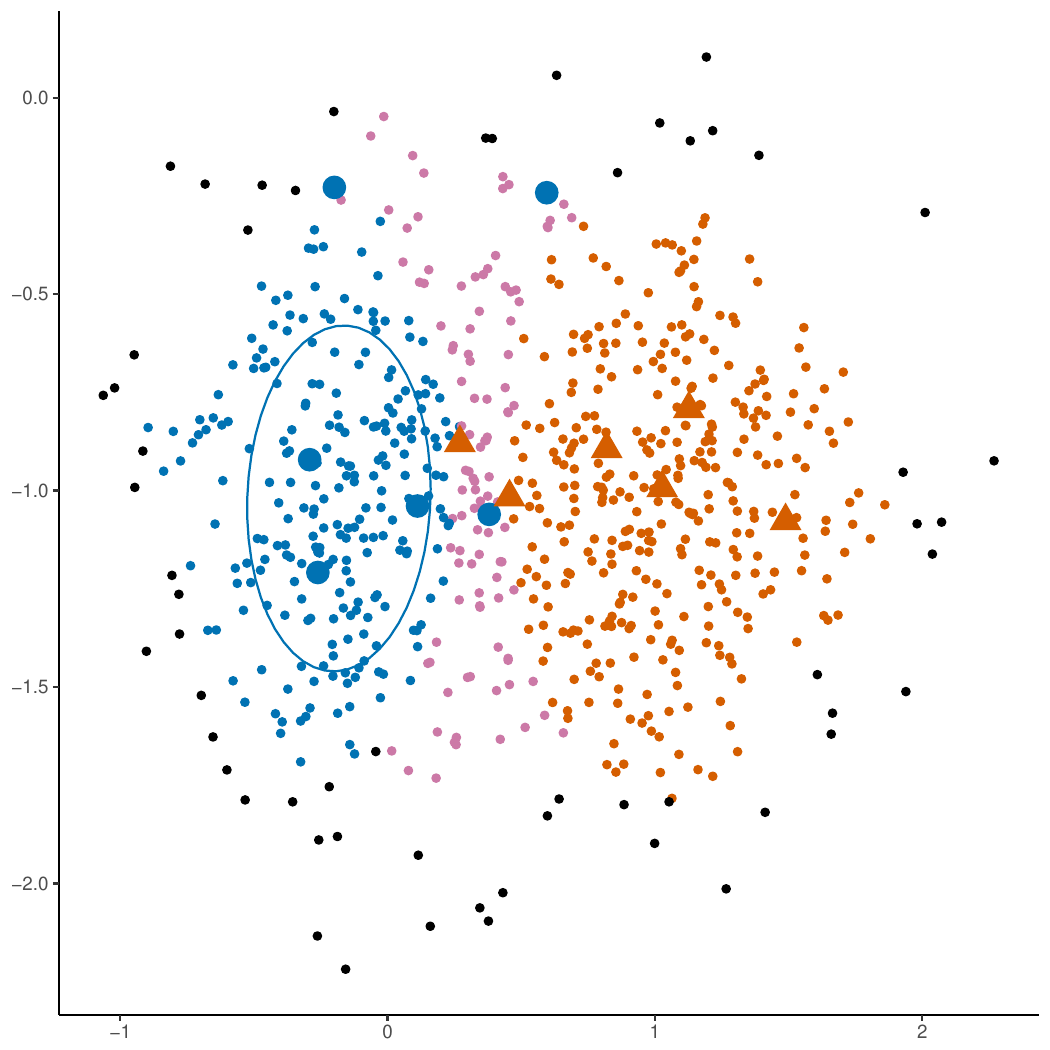}
\hskip 0.03in
\includegraphics[width=1.3in]{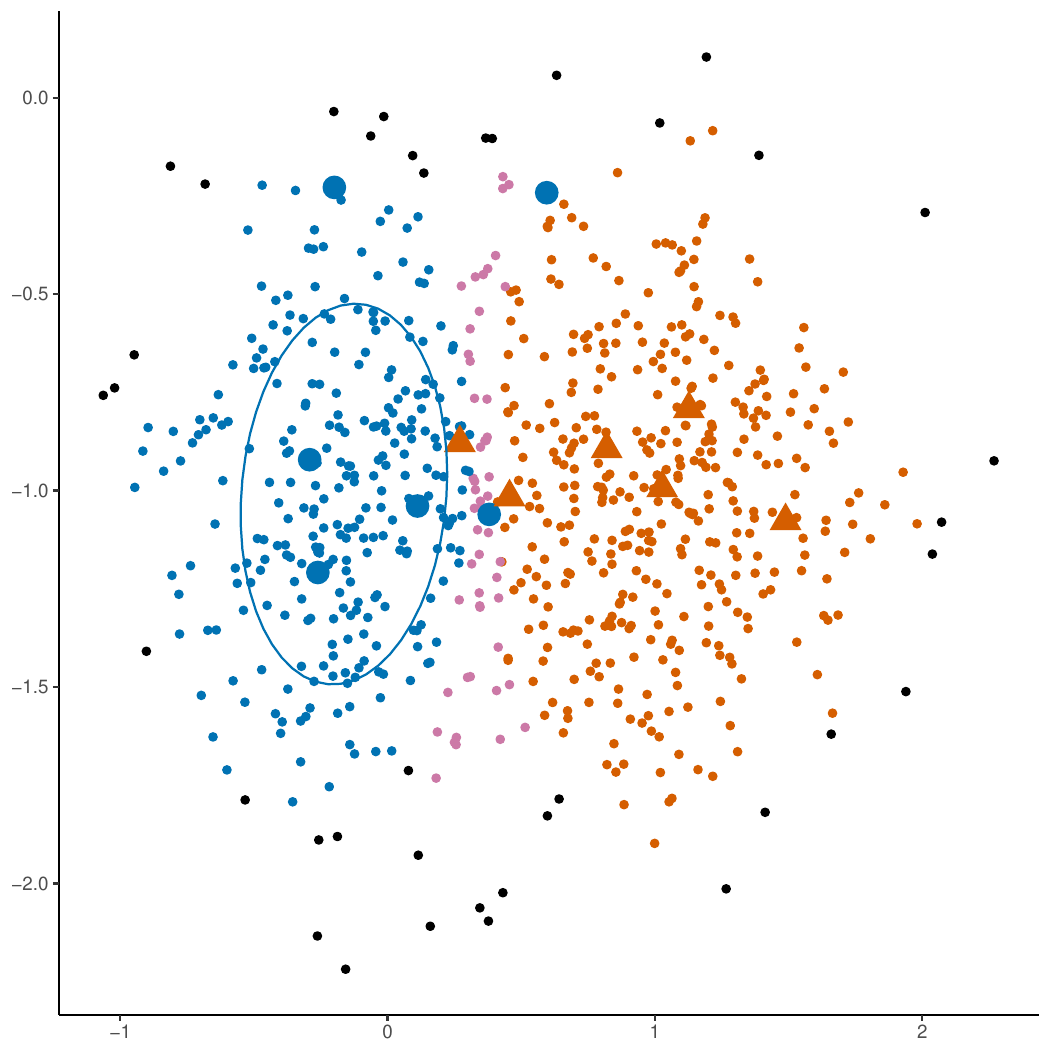}
\hskip 0.03in
\includegraphics[width=1.3in]{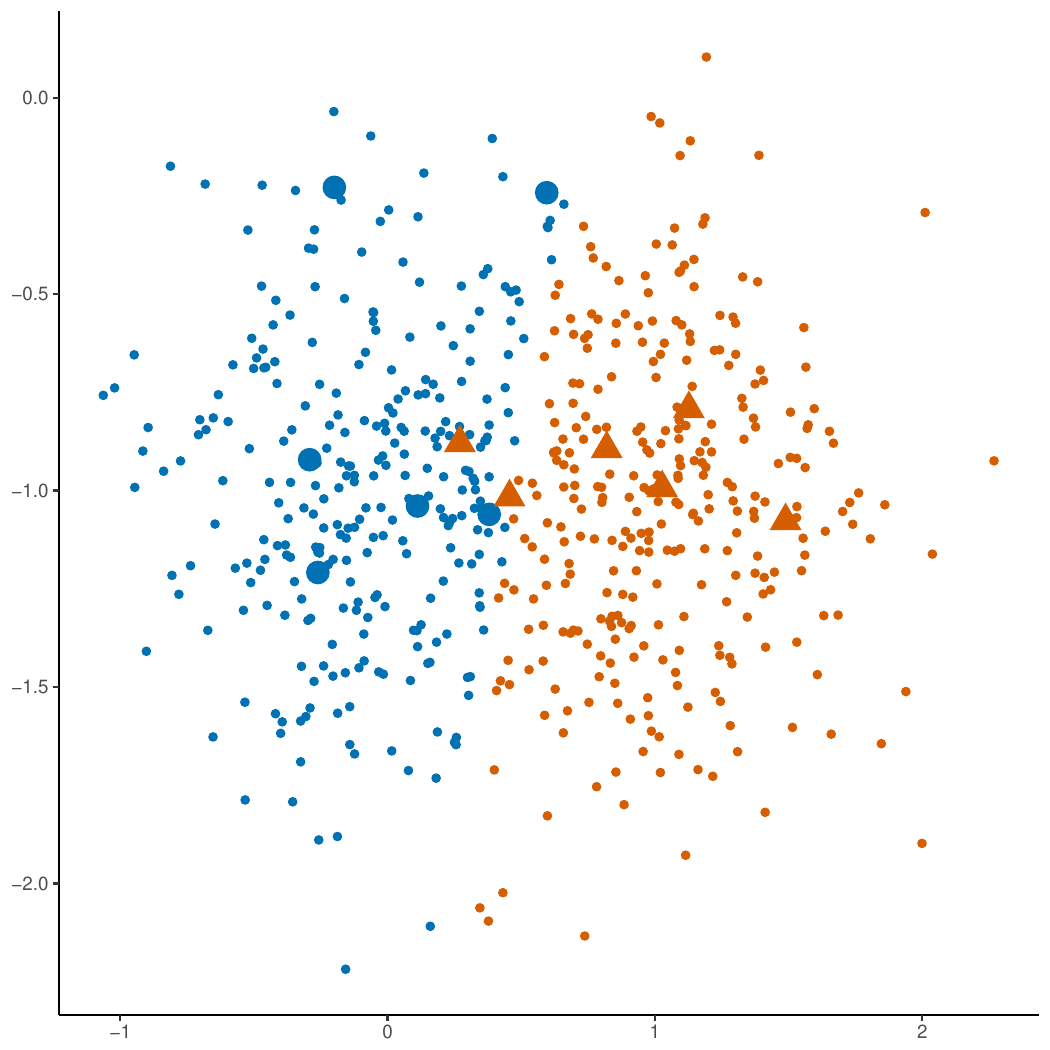}
\hskip 0.03in
\includegraphics[width=1.3in]{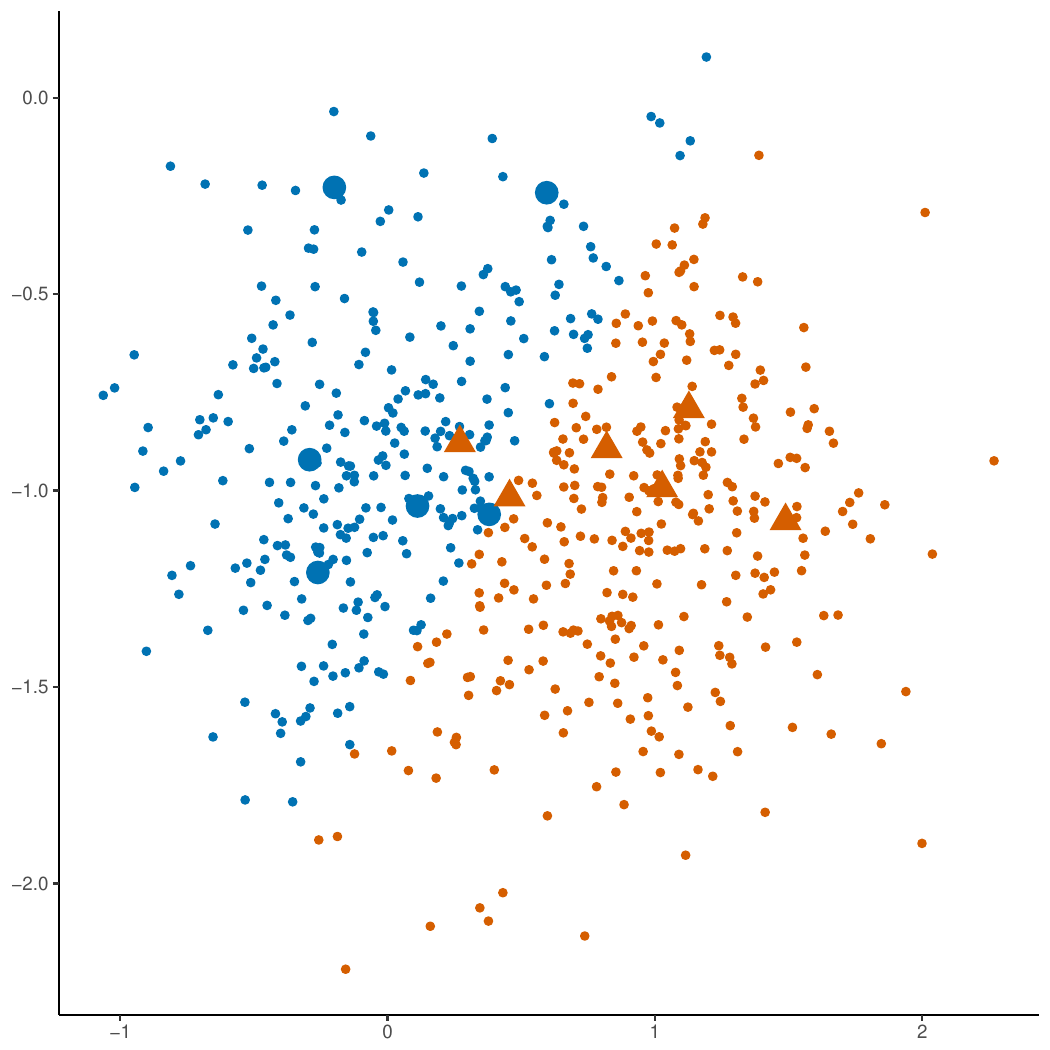}
}
\caption{Simulation 1 comparison with $\alpha=0.6$. Left to
  right: 1) ADClust with $k=10$; 2) 
  ADClust with $k=20$; 3) EM least square; 4) S4VM}  
\label{fig:sim-mixed}
\end{figure*}

\begin{figure*}[tb!]
\centering{
\includegraphics[width=1.3in]{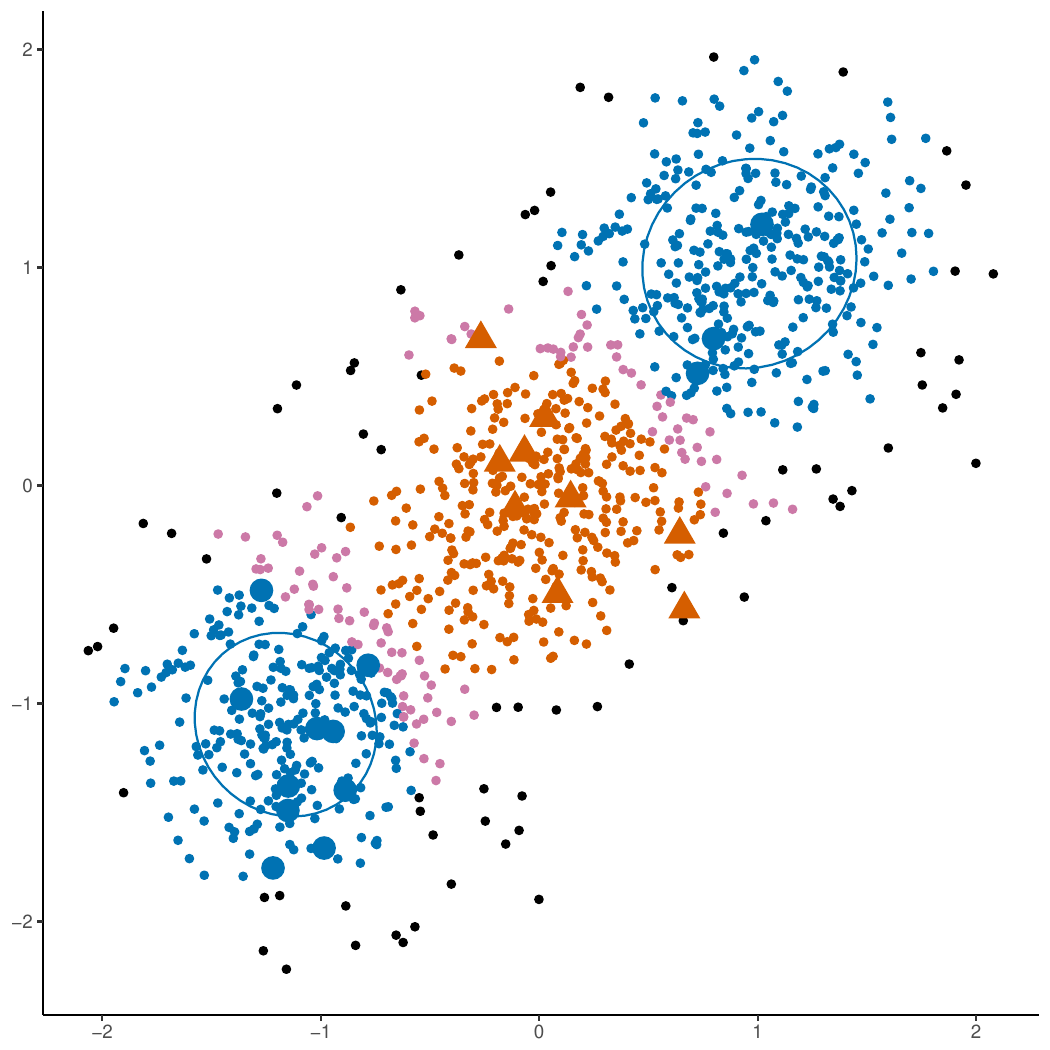}
\hskip 0.03in
\includegraphics[width=1.3in]{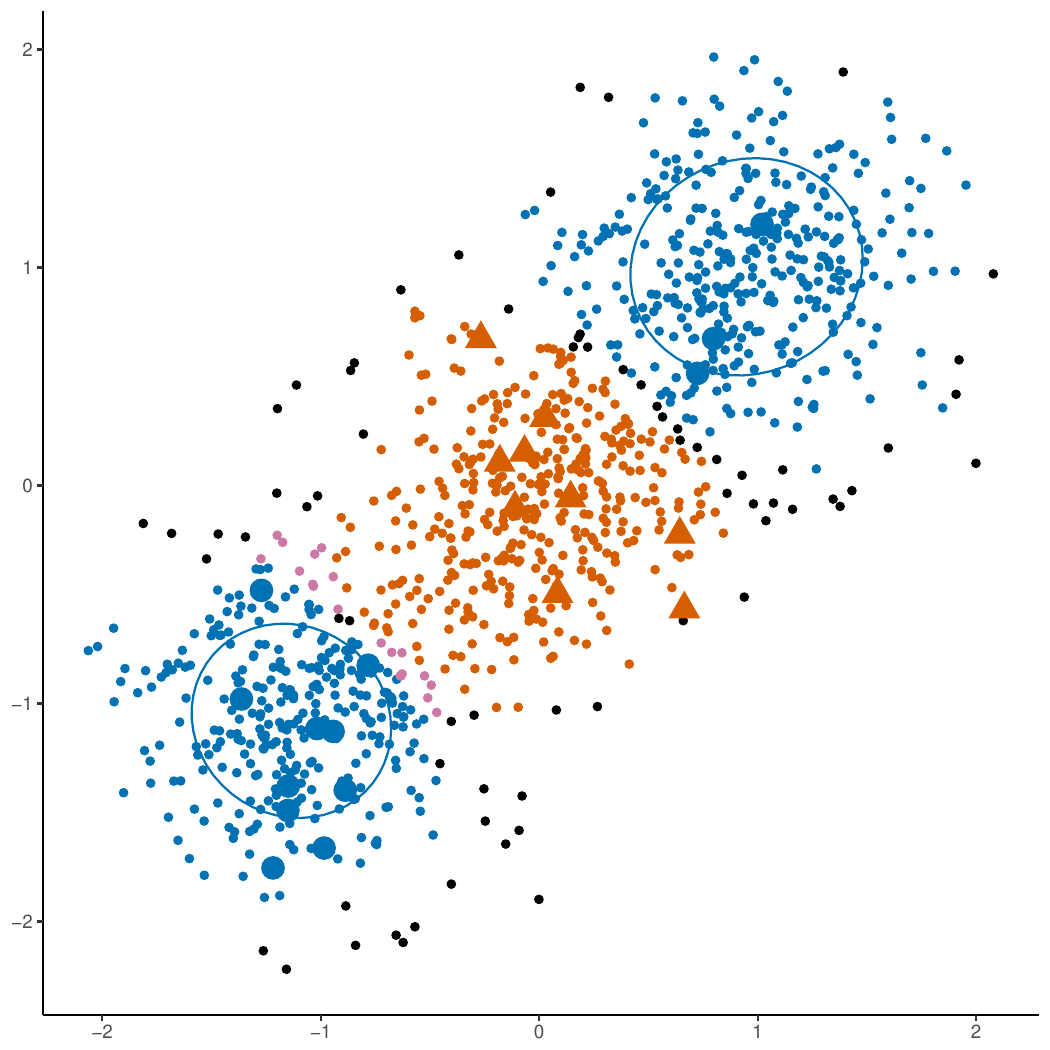}
\hskip 0.03in
\includegraphics[width=1.3in]{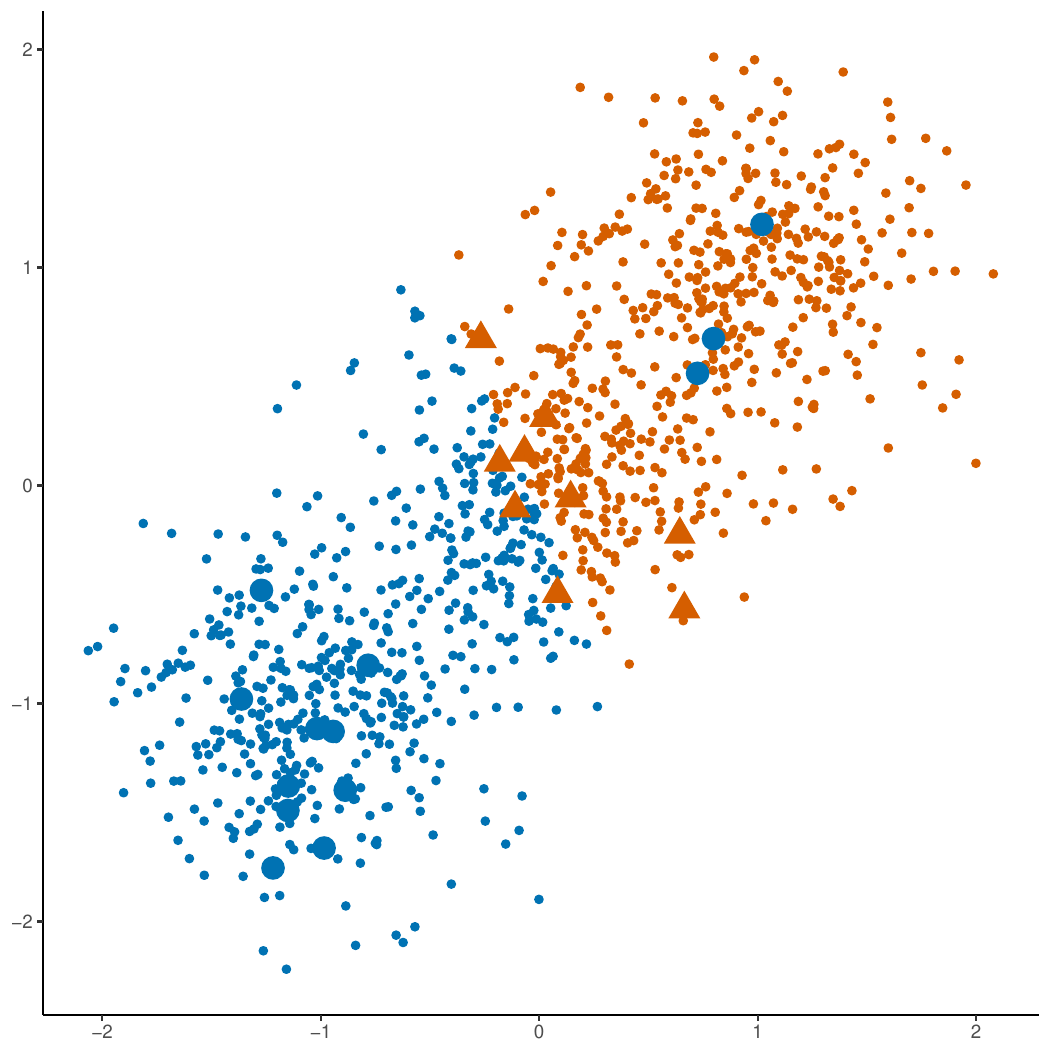}
\hskip 0.03in
\includegraphics[width=1.3in]{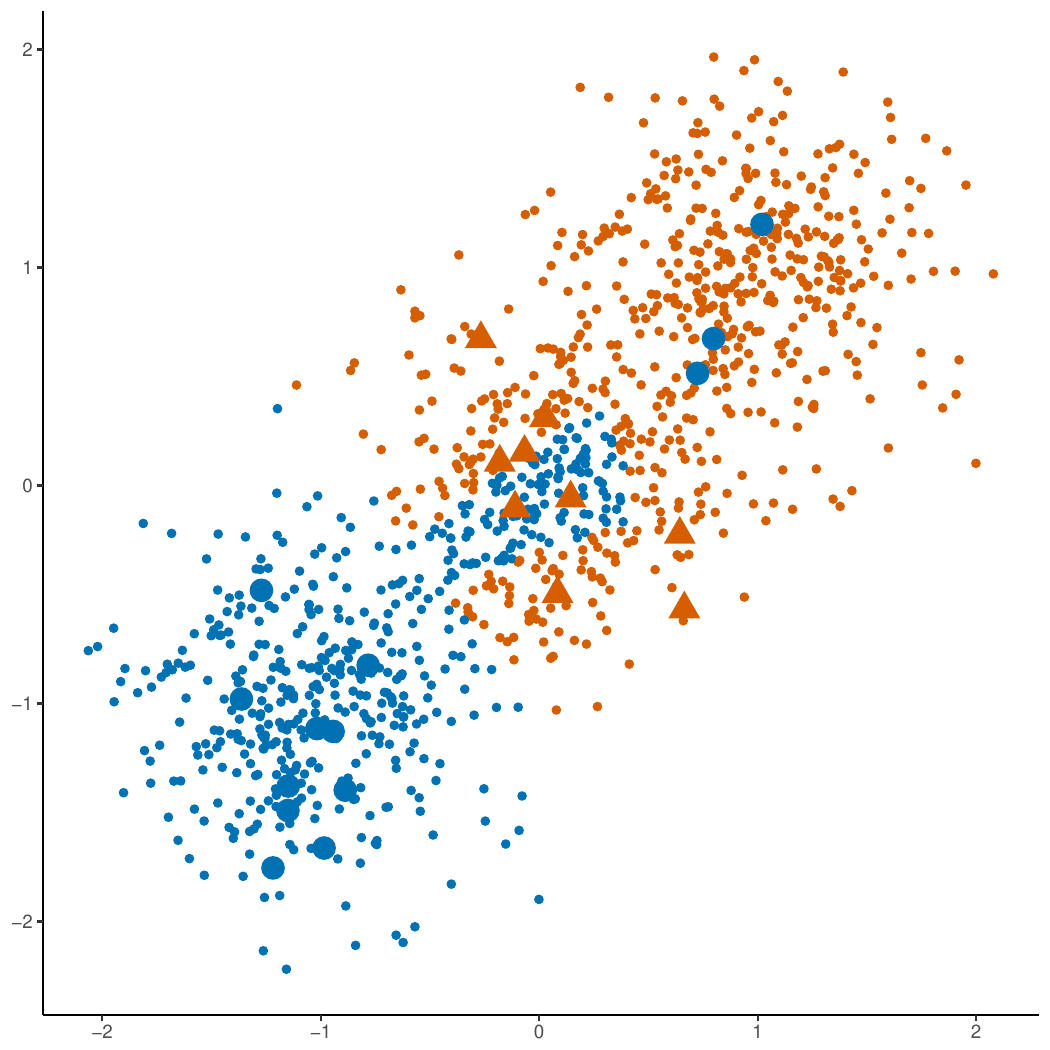}
}
\caption{Simulation 2 comparison with $\alpha=0.6$. Left to
  right: 1) ADClust with $k=10$; 2) 
  ADClust with $k=20$; 3) EM least square; 4) S4VM}  
\label{fig:sim-ABA}
\end{figure*}

\begin{figure*}[tb!]
\centering{
\includegraphics[width=1.3in]{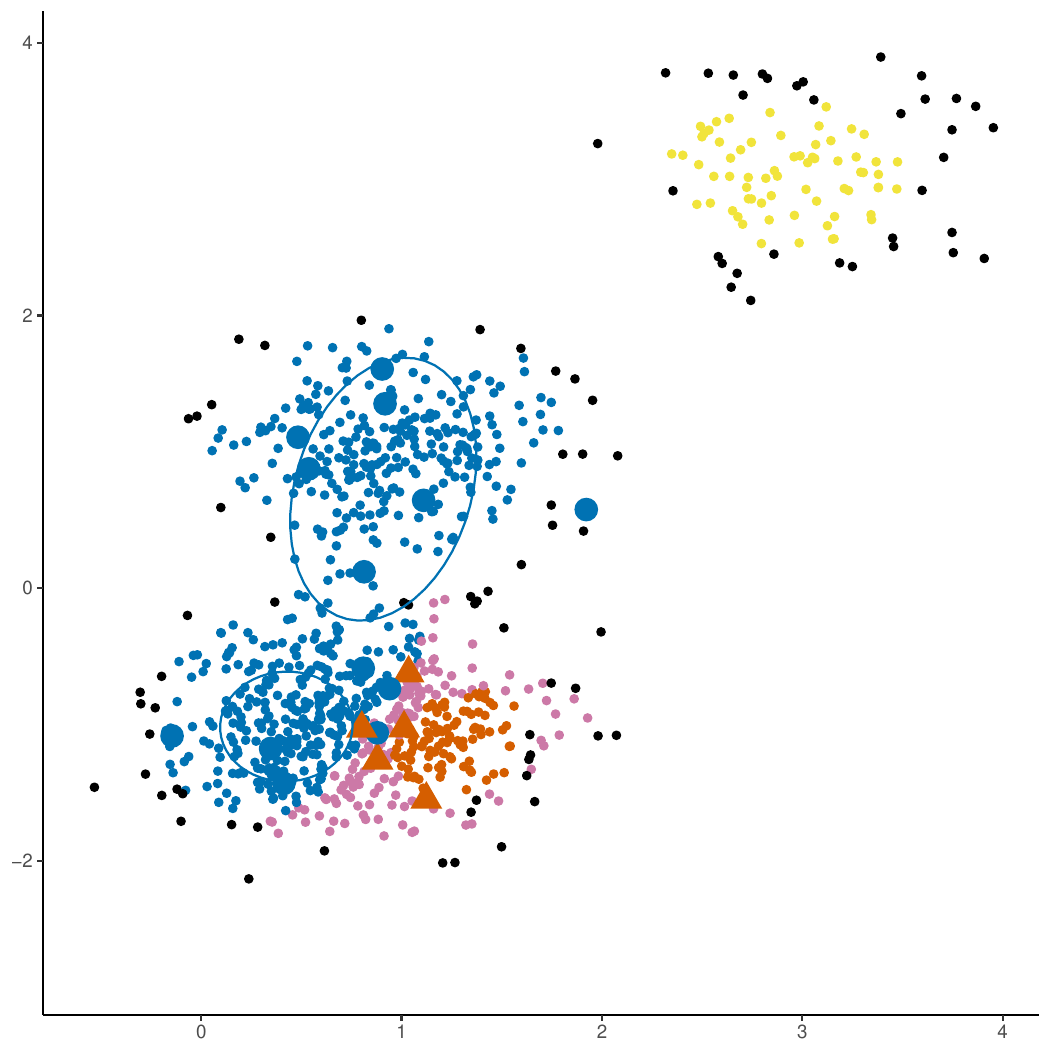}
\hskip 0.03in
\includegraphics[width=1.3in]{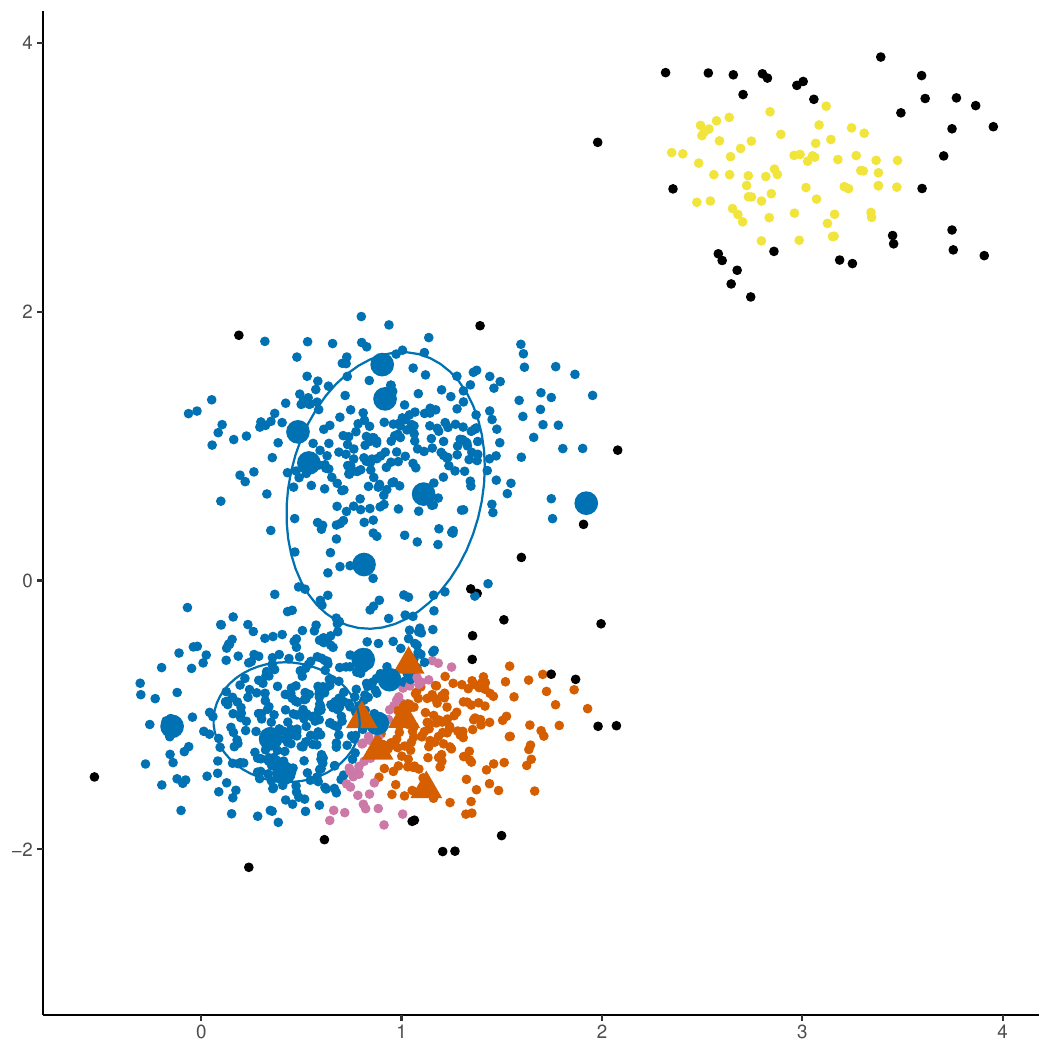}
\hskip 0.03in
\includegraphics[width=1.3in]{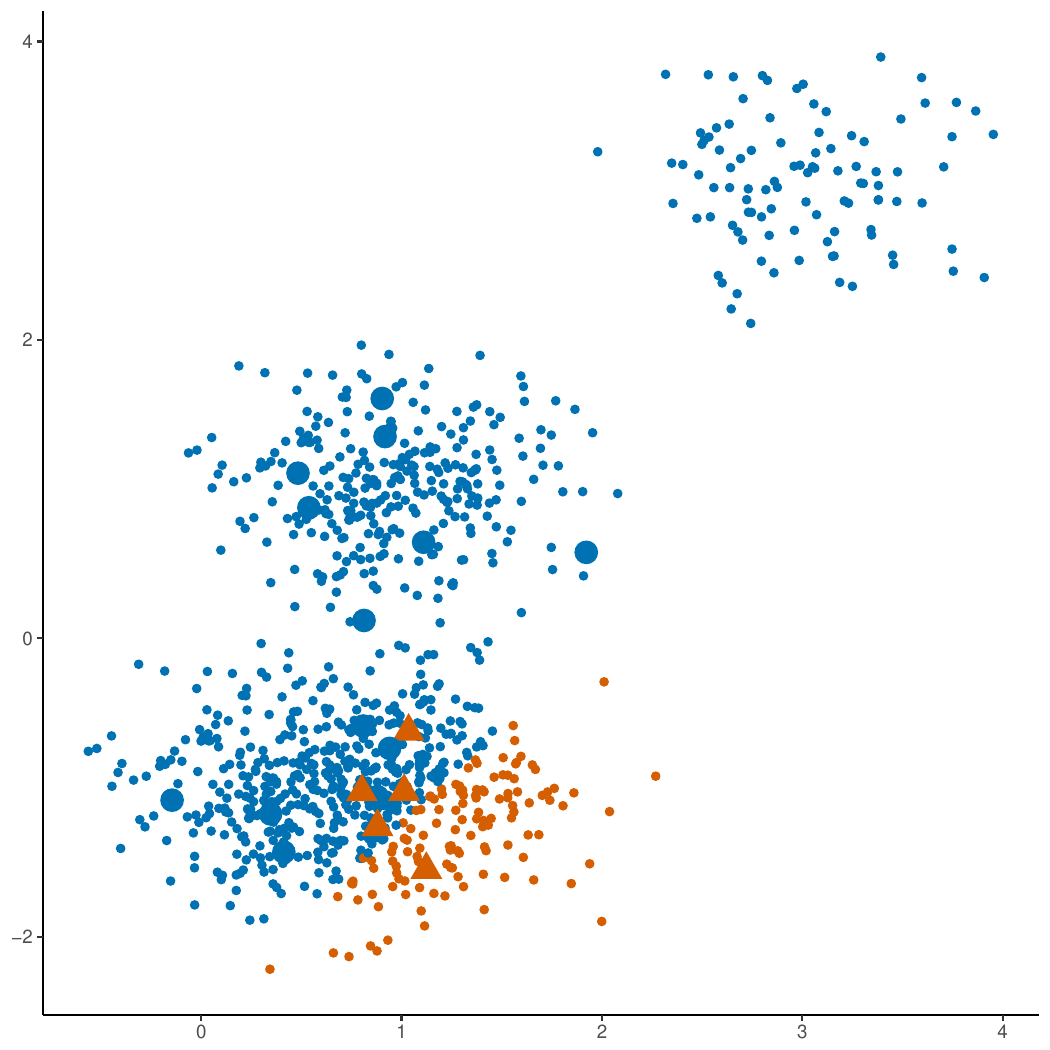}
\hskip 0.03in
\includegraphics[width=1.3in]{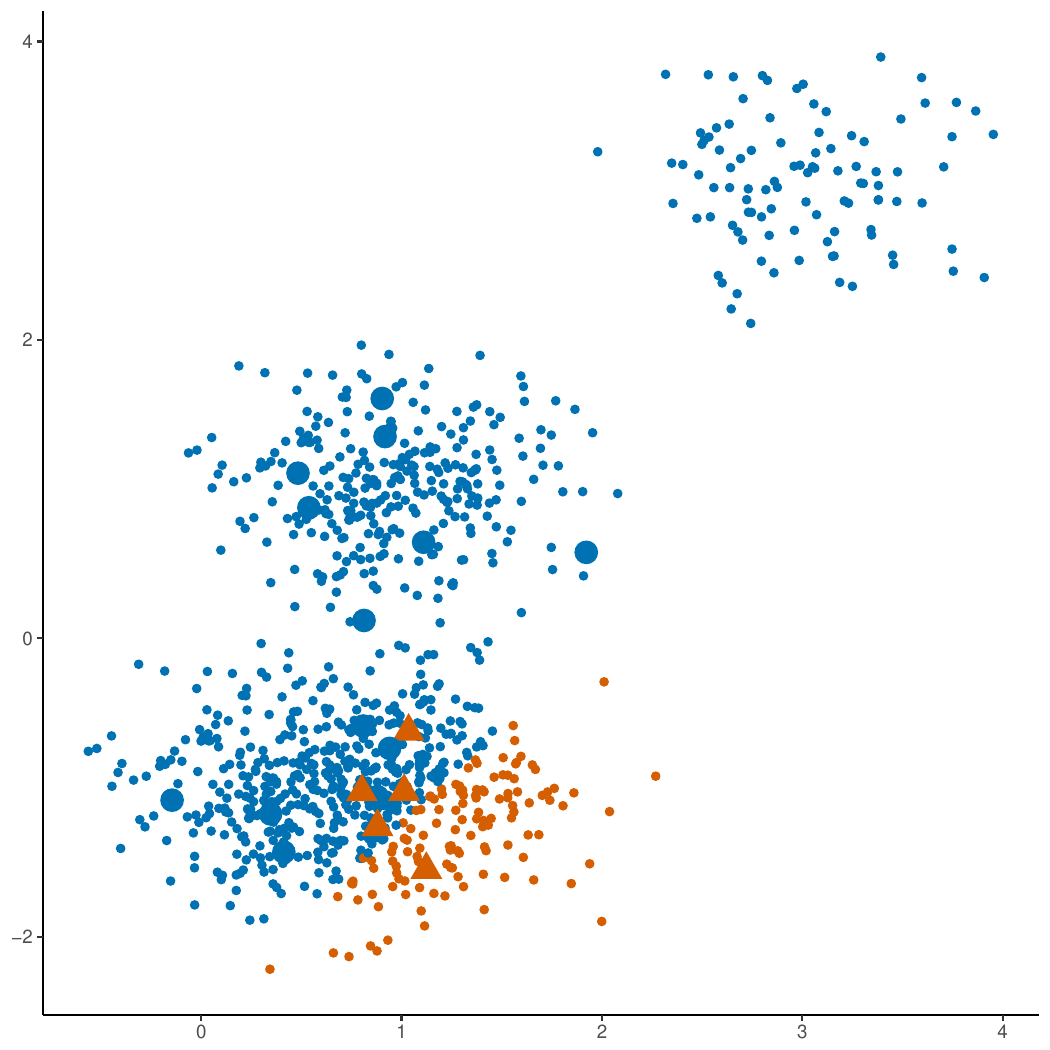}
}
\caption{Simulation 3 comparison with $\alpha=0.6$. Left to
  right: 1) ADClust with $k=10$; 2) 
  ADClust with $k=20$; 3) EM least square; 4) S4VM}  
\label{fig:sim-Unknown}
\end{figure*}

Since cyber security applications often produce big data sets, we
need a computationally efficient algorithm, which needs to be easy to tune
as well. Inspired by a traditional grid based clustering
algorithm~\cite{zhao-2001}, we develop a grid based adversarial 
clustering algorithm (ADClust). Our algorithm applies a
Gaussian kernel classifier to compute the probability scores for
every unlabeled data points. Then using a pre-specified weight ,we obtain
re-weighted density of the data points. In the first pass, our
algorithm groups the data points into normal sub-clusters,
abnormal sub-clusters, unlabeled sub-clusters and unlabeled
outliers. Notice that the choice of the
weight affects the size of the overlapping areas and the normal
and abnormal regions. Then
in a second pass, we do not use label information, and simply
group the data points into large unlabeled clusters and identify
unlabeled outliers. The next step is to match the normal,
abnormal, unlabeled smaller clusters from the first pass with the
unlabeled larger clusters from the second pass. This way we are
able to identify normal and abnormal regions within one cluster
along with the unlabeled overlapping regions. The last step is to
play a conservative strategy, drawing defensive walls inside
normal regions to ensure that we identify relatively pure normal core
positions. Figures~\ref{fig:sim-mixed}, \ref{fig:sim-ABA}, and
\ref{fig:sim-Unknown} in Section~\ref{sec:exp} show how our
algorithm work on three simulated datasets.  

During the initialization
stage of our algorithm we create the cells, compute the distance threshold RT
and the density threshold DT. We choose a pre-determined positive
weight $k$ to assign re-weighted density to every unlabeled
point. The value of $k$
affects the size of the overlapping
regions. Section~\ref{sec:exp} examine different $k$ values and
recommend $k$ around 30. $coefRT$ and $coefDT$ are also tuning
parameters. In Section~\ref{sec:exp}, we set $coefRT=20$ and $coefDT=0.95$,
which achieve good results.  
There are three initialization steps.  \\

\texttt{Initialization Step 1.} Creating cells: For every variable $X_i$, 
divide
its range [min$(X_i)$, max$(X_i)$] into $m_i$ equal sized
sections, $i=1,...,q$. We choose the number $m_i$ to ensure each section has
roughly $5\%$ to $10\%$ of the data points. For different
variables, the number of sections $m_i$ can be different. 
Hence in the $q-$dimensional space, the sections along each
dimension together form small $q-$dimensional cells. Given a particular
cell, we call the cells in its hypercube neighborhood with
radius 1 as its neighbor cells. \\    

\texttt{Initialization Step 2.} Thresholding: Compute the distance threshold RT
and the density threshold DT as follows. 
\begin{itemize}
\item Distance Threshold RT: For a data point $p$ in cell $c$, we
compute the pairwise distances $\textrm{d}(p,o)$ between $p$
and all the points in cell $c$'s neighbor cells. 
For point $p$, let $a(p)=\textrm{mean}(\textrm{d}(p,o))$ be the average of all
the pairwise distances. 
Let
$d(c)=\textrm{mean}_{p \in c}(a(p))$ be the average over all the points in
cell $c$. The distance threshold 
$$ RT = \frac{mean(d(c))}{q \times coefRT}. $$ 

\item Density Threshold DT: For a data point $p$ in cell $c$, its
  density $n(p)$ is the number of points within the distance
  threshold RT from the data point $p$. A cell $c$'s density is
  $n(c)=\textrm{mean}_{p \in c}(n(p))$, the average of the densities of the
  points in cell $c$. The density threshold 
$$DT = \frac{mean(n(c))}{\ln(N)} \times coefDT,$$
where $N$ is the total number of data points. \\  
\end{itemize}

\texttt{Initialization Step 3.} Weighting: We build a Gaussian kernel classifier with
   the handful labeled data points. Normal objects are labeled as
   1s and abnormal as 0s. We then apply the Gaussian kernel
   classifier to the unlabeled objects. Each unlabeled points
   $p$ is assigned a probability score $b_p \in [0,1]$. A
   pre-determined positive weight 
   $k$ is used to map the scores $b_p$ from [0,1] to [-$k$,$k$].  \\

Our adversarial clustering algorithm has five steps. Algorithm 1
shows the function Merge. Algorithm 2 is the main algorithm, with
the initialization steps and the following five steps.\\ 

\texttt{Merge 1:} Creating labeled normal and abnormal
sub-clusters: Use each point's re-weighted density $n(p)\times b_p$.  
First take the points whose
re-weighted densities are 
   greater than density threshold DT as cluster centroids. Merge
   the remaining points with the cluster centroids if their
   distances to a cluster centroid is less than distance
   threshold RT. If a point's distance to multiple cluster
   centroids are less than RT, then those small clusters are
   merge into one big cluster. Continue to merge. The data points
   not assigned to any cluster remain unlabeled. \\ 
 
\texttt{Merge 2:} Clustering the remaining unlabeled data points:
   Remove all the normal
   and abnormal sub-clusters. For the remaining data points,
   use their original density $n(p)$. Merge the remaining
   unlabeled data points.  \\

\texttt{Merge 3:} Using the same $RT$, $DT$, and $k$ parameter
values, and 
every data point's original density $n(p)$, we merge all the
data points without considering the labels. We obtain unlabeled
clusters, and unlabeled outliers. \\

\texttt{Match:} Match the above unlabeled clusters
with the normal and abnormal sub-clusters, and the clusters of the
remaining unlabeled data points from the first pass. 
Now we are able
to identify clusters 
which contain normal and abnormal regions and
their overlapping areas. The points in the overlapping areas are
not labeled. If there are remaining unlabeled clusters,
they are outlying unknown clusters. The rest are outliers, i.e., potential
anomalies. \\  

\texttt{Draw defensive walls:} We draw $\alpha$-level defensive walls
inside the normal regions to ensure that we protect the relatively
pure normal positions. \\

\begin{algorithm}
 \KwData{d(p), n(p), DT, RT}
% \KwResult{The clusters}
 \caption{Function Merge}
 \For{point $p_j$ in the space}{
 	\If{n($p_j$) $\geq$ DT}{
 		Assign $p_j$ a new label\;
 	}
 }
 \While{No more clusters can merge}{
 	\For{every cluster $cl_i$}{
 		\For{every cluster $cl_j$}{
 			\If{exist a point $p_1$ in $cl_i$ and a point $p_2$ in $cl_j$, distance($p_1$,$p_2$) $\leq$ RT}{
 				Merge($cl_i$, $cl_j$)\;
 			}
 		}
 	}
 }
\end{algorithm}

\begin{algorithm}
 \KwData{Unlabeled points, labeled points, confidence $\alpha$}
% \KwResult{The predicted labels}
 initialization\;
 q $\gets$ dim(space)\;
 N $\gets$ count(points)\;
 k $\gets$ function($\alpha$)\;
 \For{every cell $c_i$ in the space}{
 	\For{every point $p_{ij}$ in the cell $c_i$}{
 		\For {every point $q_{ijk}$ in $c_i$'s neighborhood}{
 			d($p_{ij}$, $p_{ijk}$) $\gets$ distance($p_{ij}$,$p_{ijk}$)\;
 		}
 		d($p_{ij}$) $\gets$ mean(d($p_{ij}$, $p_{ijk}$)) for every k\;
 	}
 	d($c_i$) $\gets$ mean($p_{ij}$) for every j\;
 }
 RT $\gets$ $\frac{Mean(d)}{q \times coefRT}$\;
 \For {every cell $c_i$ in the space}{
 	\For {every point $p_{ij}$ in the cell $c_i$}{
 		n($p_{ij}$) $\gets$ count(distance($p_{ij}$, $p_{ijk}$) $\leq$ RT) where $p_{ijk}$ are the points in $c_i$'s neighborhood\;
 	}
 	n($c_i$) $\gets$ mean($p_{ij}$) for every i\;
 }
 DT $\gets$ $\frac{mean(n(c_i))}{lg(N)} \times coefDT$\;
 \For {every point $p_j$ in the space}{
 	$w(p_j)$ $\gets$ k $\times$ GaussianKernalClassifier($p_j$, points, known labels) \;
 }
 Assign(abnormal or normal) $\gets$ Merge(d(p) * w(p), n(p), DT, RT)\;
 Assign(no-label to remaining points) $\gets$ Merge(d(p neither
 abnormal nor normal), n(p neither abnormal nor normal), DT, RT)\;
 Assign(no-label to all points) $\gets$ Merge(d(p), n(p), DT,
 RT)\;
 Match Assign(abnormal or normal) and Assign(no-label to
 remaining points) with Assign(no-label to all points)\; 
 Draw $\alpha$-level defensive walls inside normal sub-clusters\; 
 \caption{Main Clustering Algorithm}
\end{algorithm}

\begin{figure*}[tb!]
\centering{
\includegraphics[width=1.6in]{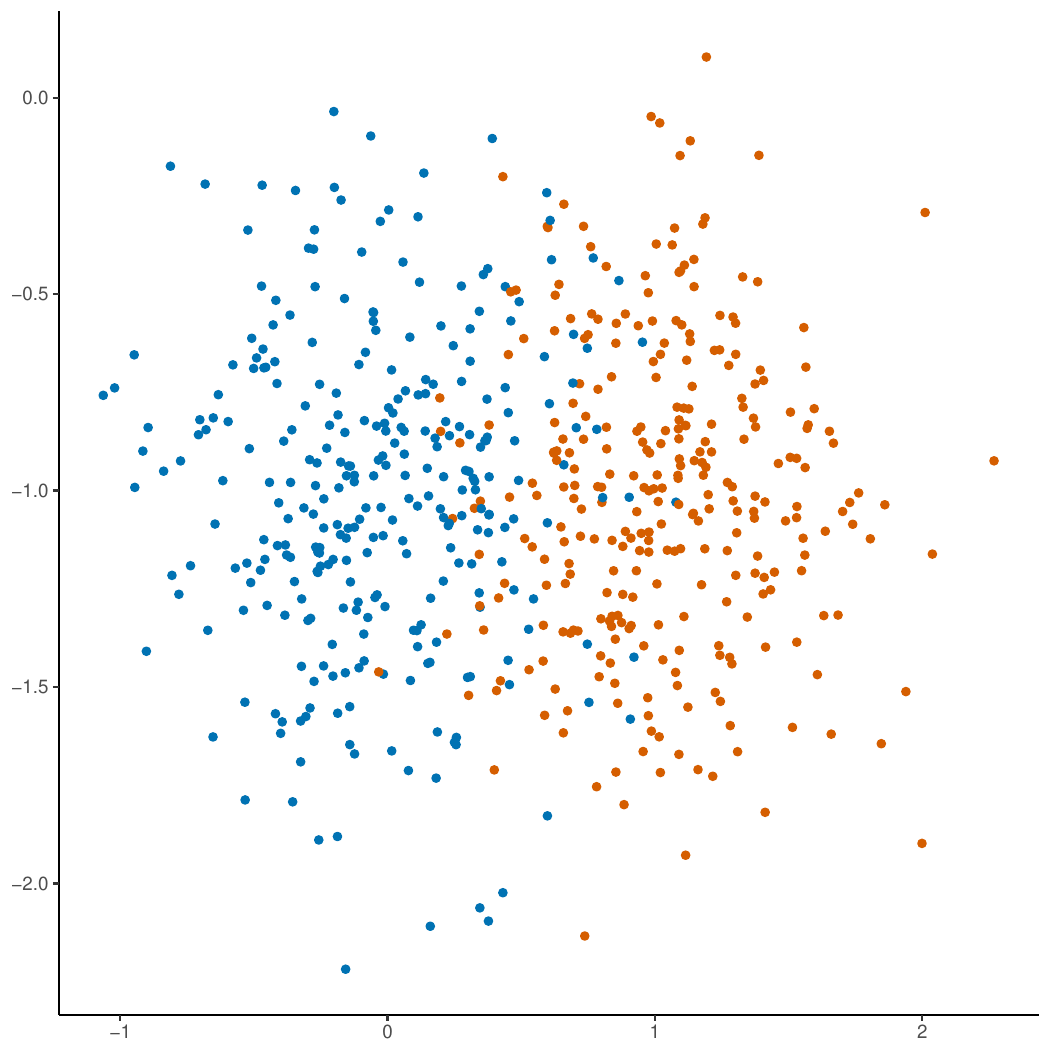}
\hskip 0.03in
\includegraphics[width=1.6in]{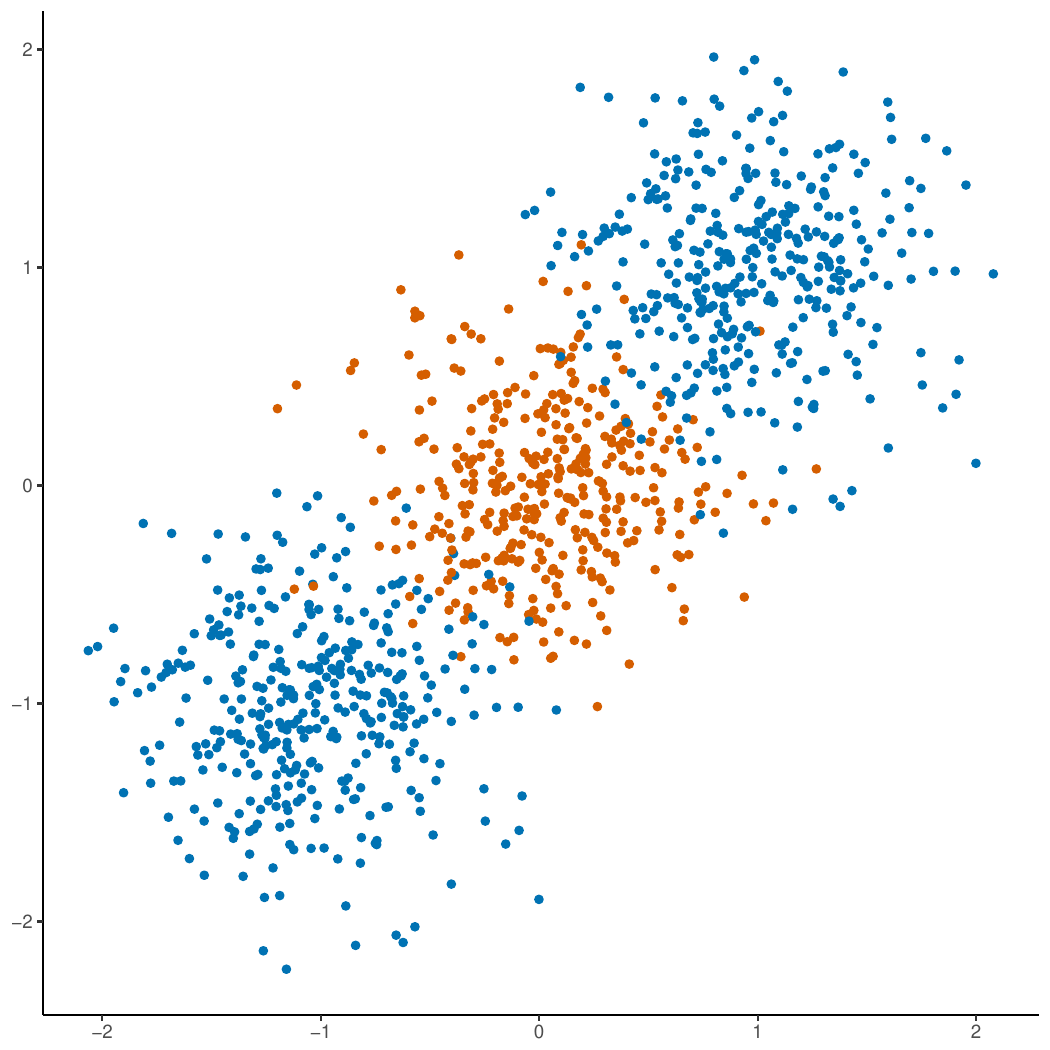}
\hskip 0.03in
\includegraphics[width=1.6in]{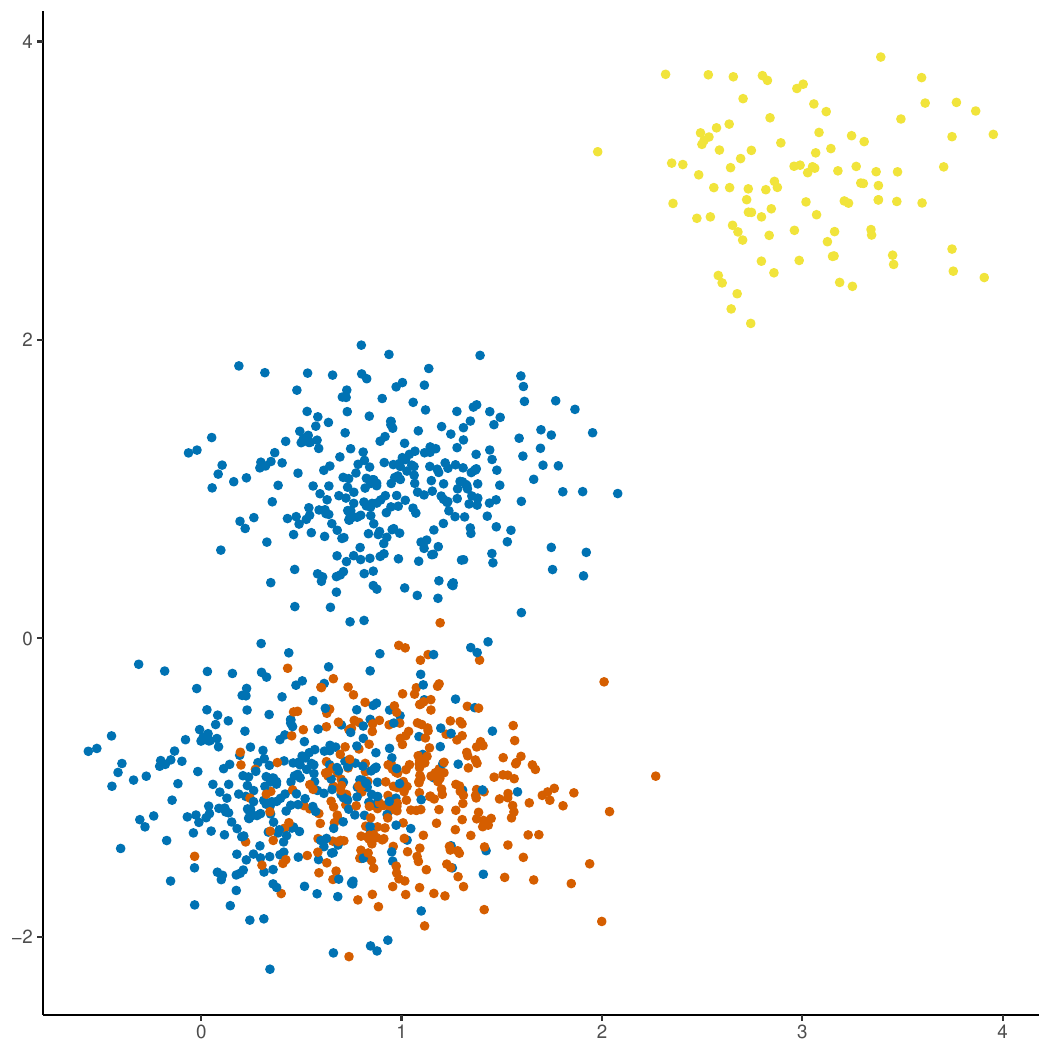}
}
\caption{Left to right: 1) Simulation 1 true labels;
  2) Simulation 2 true labels; 3) Simulation 3 true labels;}  
\label{fig:sim-true}
\end{figure*}

\begin{figure*}[tb!]
\centering{
\includegraphics[width=2.35in]{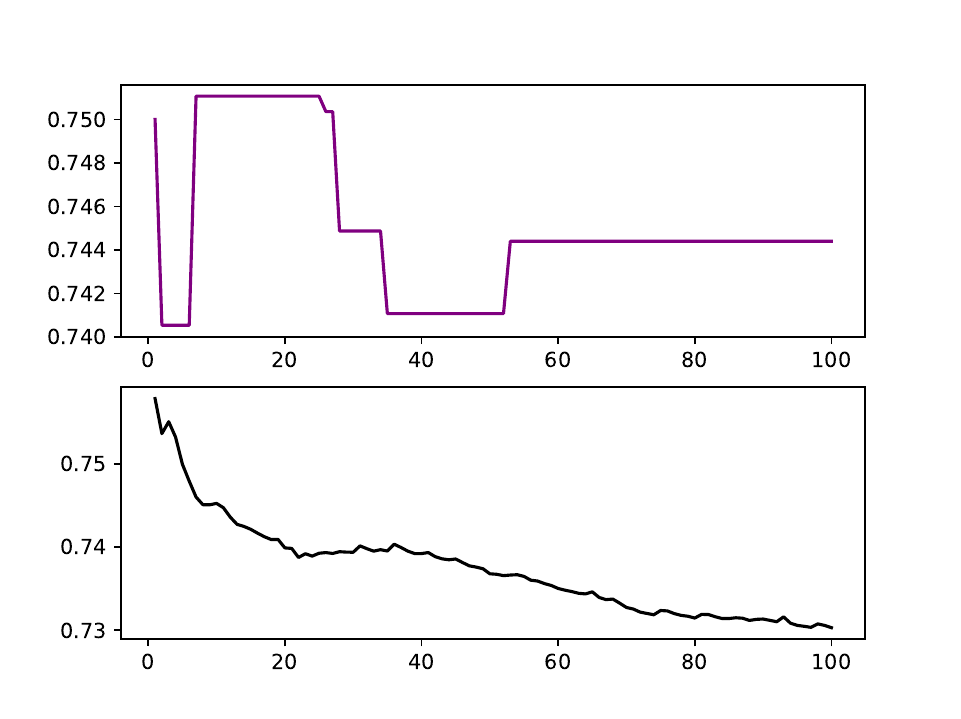}
\hskip 0.03in
\includegraphics[width=2.35in]{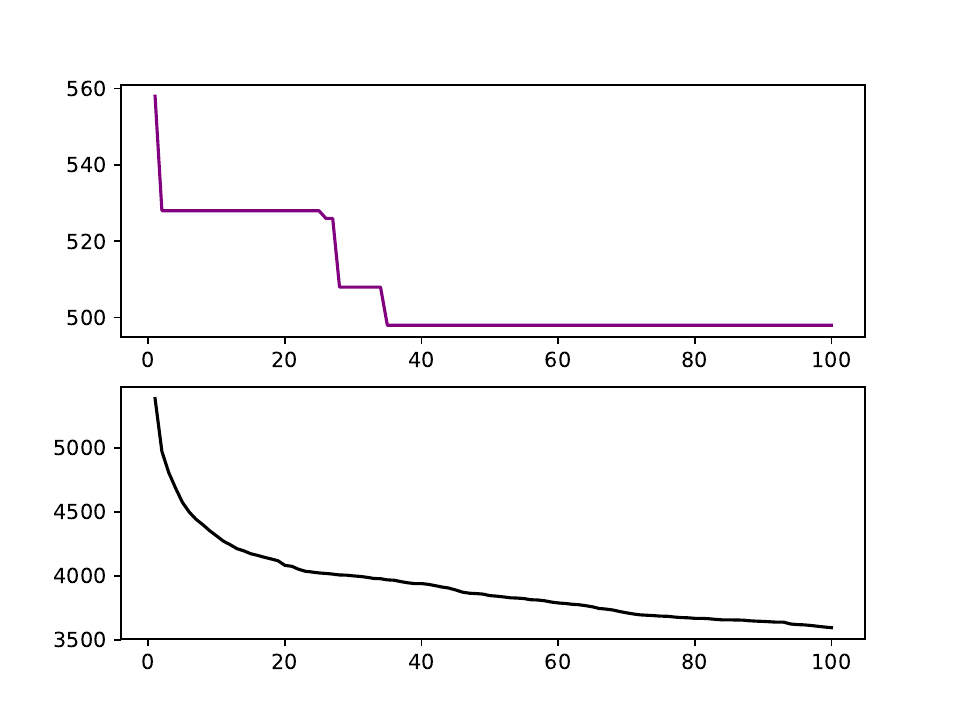}
}
\caption{Quantitative measures as the weight $k$
  increases from 1 to 100. 
%Left Column: Top panel is the
%  pureness rate of abnormal clusters and bottom panel is the
%  pureness rate of normal clusters;   
  Left Column: Top panel is percent of abnormal points in mixed
  region and bottom 
  panel is percent of abnormal points among outliers; Right Column: Top panel
  is the number of 
  points in mixed region and bottom panel is the number of points
  as outliers.} 
\label{fig:pureness}
\end{figure*}

\begin{figure*}[tb!]
\centering{
\includegraphics[width=3.35in]{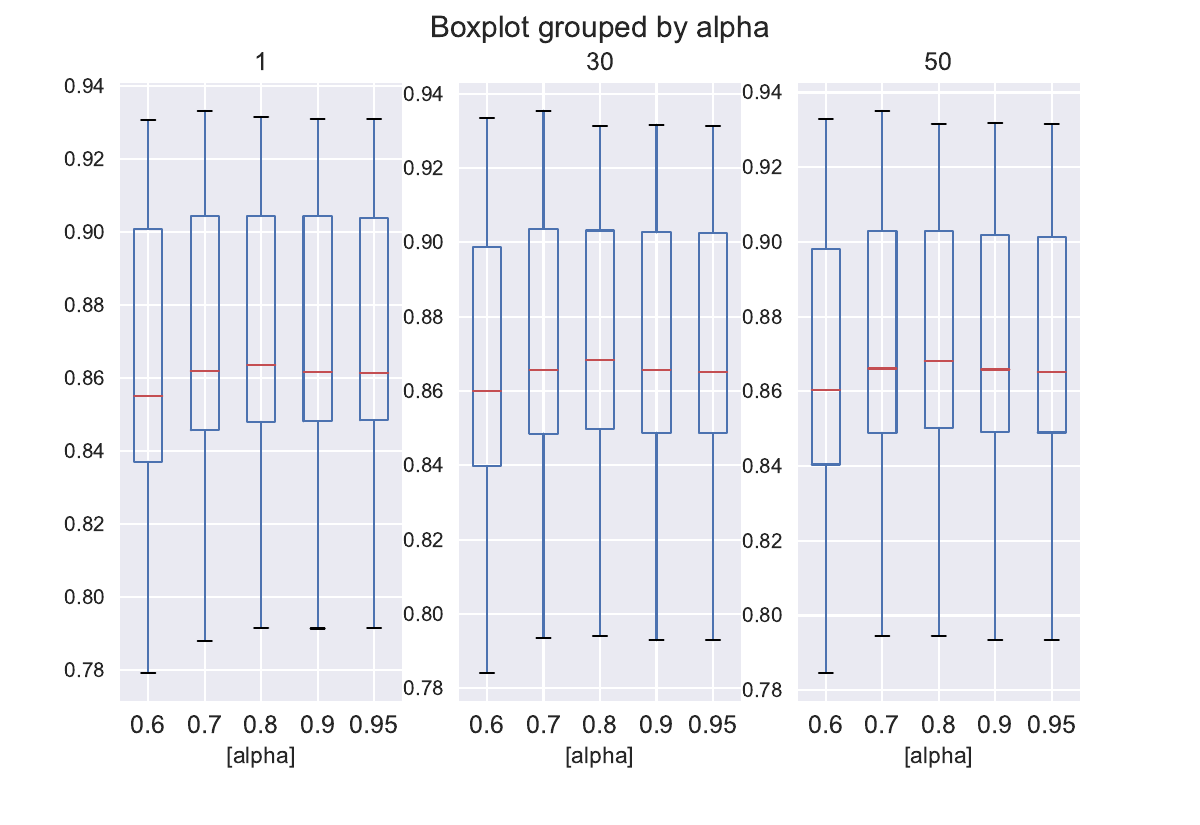}
\hskip 0.05in
\includegraphics[width=3.35in]{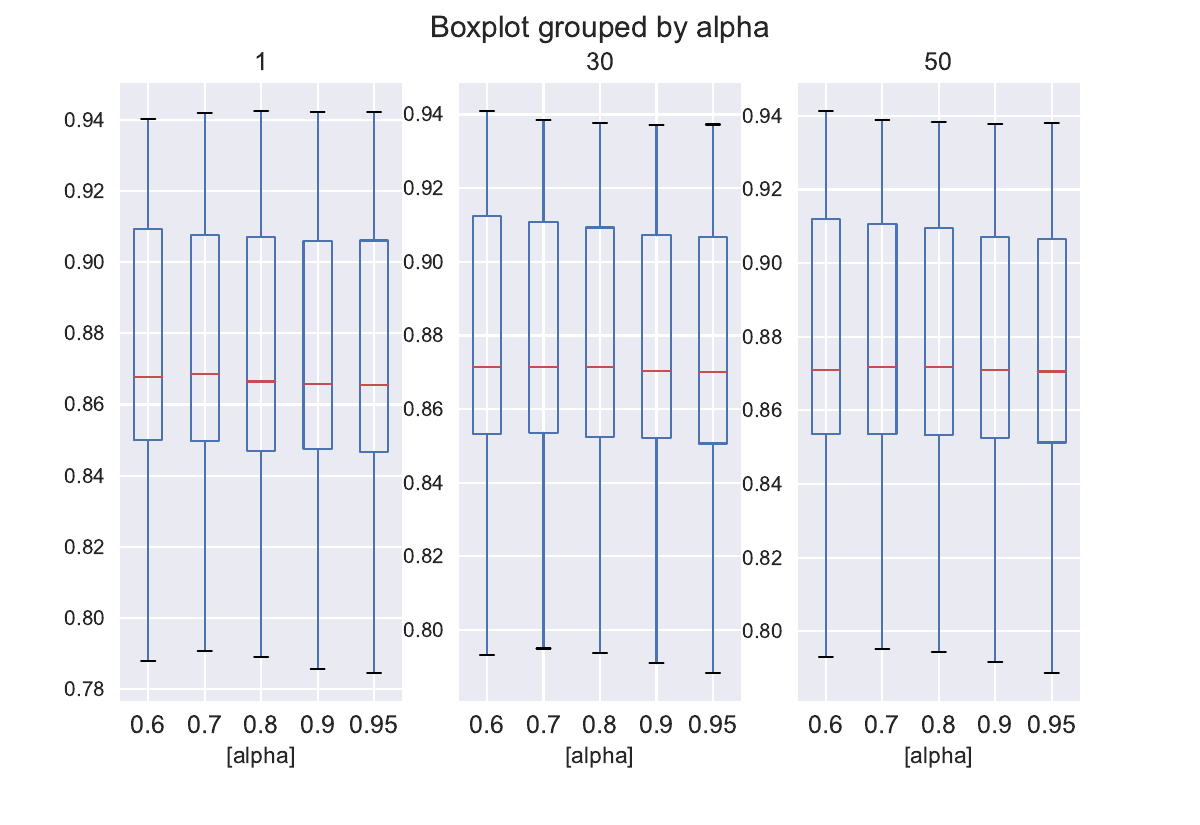}
}
\caption{Left Panel:
  The boxplots of percent of normal objects within Manhattan
  defensive walls with 
  different $\alpha$ levels and 
  weight $k$. The corresponding $\eta(\alpha)$ values are 5.89, 7.25, 8.97,
  11.51, 13.72; Right Panel: The boxplots of percent of normal
  objects within Euclidean defensive walls 
  with different $\alpha$ levels and weight scalars.} 
\label{fig:boxplot}
\end{figure*}

\section{Experiments}
\label{sec:exp}

\subsection{Simulated Experiment}

We conduct three simulations to compare our ADClust with two
semi-supervised learning algorithms,  EM least square \cite{compare-em} and
S4VM \cite{compare-svm}. In all three simulations, we generate data points
from several bivariate normal distributions. The bivariate normal
distributions have the same variance-covariance matrix
$$
\Sigma=\begin{bmatrix} 0.4 & 0 \\ 0 & 0.4 \end{bmatrix}. 
$$ 

Figure~\ref{fig:sim-true} shows the true labels of the points for the three
simulations. Solid triangles and solid circles are the 2\%
labeled points. Figures~\ref{fig:sim-mixed}, \ref{fig:sim-ABA}, and
\ref{fig:sim-Unknown} show the comparison results. We set
$\alpha=0.6$. Blue dots are
used for normal points, either true or labeled. Orange dots are used
for abnormal points,  either true or labeled. Purple dots are for
unlabeled points in mixed regions. Yellow dots are for unknown
unlabeled clusters. Black dots are for unlabeled outliers. Different
regions are marked using different colors.  

In these three simulations, we show the data points after attacks
have taken place. In Simulation 1 previously separated normal and abnormal
clusters now have overlapped areas and are merged into one big
cluster. In Simulation 2, an attack
takes place between two normal regions, and attack objects manage
to mix three clusters into one big cluster.  
Simulation 3 suffers the strongest attack, where normal and abnormal
clusters are heavily mixed. There is also a previously unknown
cluster in simulation 3, which cannot be
identified as either normal or 
abnormal at the training time, potentially a new attack. 

Defensive walls, studied under a game theoretic framework, are a
crucial factor in our algorithm. With 
wall size $\alpha=0.6$, as in the recommended range from the game theoretic
study considering defender being a follower, the defensive walls mark out
the center areas of the normal regions with nearly pure normal
objects in Simulation 1 and 2, where attacks have not yet reached 
the centers of the normal. In Simulation 2, the defensive
walls in our ADClust algorithm 
successfully mark out the two centers of the two normal
regions. On the other hand the semi-supervised learning
algorithms still make a hard separation of normal
vs. abnormal. One normal cluster is completely wrongly labeled by
the two semi-supervised learning algorithms. 
In Simulation 3, our algorithm leaves an previously unknown
cluster unlabeled. It needs to be examined carefully later since it
can potentially be a new attack. The two semi-supervised learning
algorithms label the unknown cluster as normal, making a high risk 
decision.    

\texttt{{\bf Simulation 1:}} There are two sets of random samples
generated from two bivariate normal 
distributions, centered at (0, -1) (normal class), and (1, -1)
(abnormal class) respectively. Each
has 300 data points. We random select $2\%$ of the points and
save their labels. EM least square and S4VM make a hard
separation, and try their best to label all the points in the
mixed region. Hence they make noticeable mistakes in the mixed
region. Our ADClust does not assign class labeled to the points
in the mixed region. We instead mark out the whole region. The
comparison between our marked mixed region and a classification
boundary is similar to a confidence band vs a point estimate. We also
identify and leave outliers unlabeled, as shown in Figure~\ref{fig:sim-mixed}. 
  %MK: the mark out regions are not clear when printed black and white . 

\texttt{{\bf Simulation 2:}} There are three sets of random
samples generated from three bivariate
normal distributions, centered at (-1, -1) (abnormal class), (0,
0) (normal class), and (1, 1) (abnormal class) respectively. Each
has 300 data points. We random select $2\%$
of the points and save their labels. EM least square and
S4VM divide the big cluster into two areas. They fail to
distinguish three overlapping clusters. Our ADClust is
able to identify the three clusters and mark out the two mixed
regions, without assigning class labels there, as shown in Figure~\ref{fig:sim-ABA}. 

\texttt{{\bf Simulation 3:}} There are four sets of random
samples generated from four bivariate
normal distributions, centered at (0.5, -1) (normal class), (1,
-1) (abnormal class), (1, 1) (normal class), and
(3, 3) (unknown class). They have 300, 300, 300, 100 data points
respectively. We 
random select $2\%$ of the points from only normal and abnormal classes
and save their labels. The unknown class has no labeled point.   
EM least square and S4VM label the unknown cluster as normal,
where there is no reliable 
information. Furthermore, their assigned labels are highly inaccurate in the 
mixed regions. Our ADClust leave out the
unknown cluster as unlabeled, and identify the mixed region
without assigning labels there, as shown in
Figure~\ref{fig:sim-Unknown}.  

In the three simulations, we use two weight values. $k=10$ and
$k=20$. A smaller weight $k$ is a more conservative strategy,
i.e., we get smaller labeled regions and a larger
unlabeled mixed region. On the other hand, if we use a larger
weight $k$, it is a 
more aggressive strategy. We expect larger labeled regions and
smaller unlabeled mixed region. Drawing defensive walls following
the game theoretic study correctly identify 
the normal centers, while semi-supervised learning algorithms
completely fail to do so under certain scenarios.  

%MK: I was expecting an attack experiment where we show that the accuracy etc. under attack similar to our ad-svm work. 
%MK: Also, from discussion how well the advclustering is doing against the semi-supervised algorithm is not clear. 
%MK: Finally, game theoretic analysis kind of lost in this section. Maybe adding a discussion saying game theoretic analysis guided our parameter selection could be useful ??
%%
\subsection{KDD Cup 1999 Data}
\label{sec:experiment-kdd99}

The KDD cup 1999 data 
was initially created by MIT Lincoln Labs \cite{kddcup99}. 
The full dataset contains
about 126k labeled objects for training purpose. Around 40 percent of the objects are
network intrusion instances. There are 41 features for each
object. \cite{kdd-selection} ranked the 41 features with respect
to their effectiveness in separating normal instances from
abnormal instances. 
We use the KDD Cup 99 data to demonstrate how our ADClust
algorithm performs. We take 
25192 instances from training set. We include top 7 continuous features
according to \cite{kdd-selection} for each instance. 

In the first experiment, in a single run, we
randomly sample 150 instances and 
keep their labels. The rest are treated as unlabeled instances in
the run. We perform 100 runs. An overwhelming
majority (i.e., 99.4\%) 
of the instances are unlabeled. 

We gradually increase the weight $k$ from 1 to 100. Along with the
increasing weight, we have less unlabeled points. As a result,
it is a more aggressive 
strategy. Meanwhile, the normal region
increases and it includes more points which are more likely to be 
mis-labeled abnormal objects. Therefore, we have a trade-off
in choosing weight $k$. It is a trade-off between
the size of the labeled regions within a cluster and the error
rate of mis-labeled points. 
In Figure~\ref{fig:pureness}, the number
of points in mixed regions 
and outliers decreases as we have a larger weight. The percentage
of abnormal objects in the mixed areas and among outliers 
decreases from 76\% to 73\% as 
the weight $k$ increases, which 
means we have to exam the mixed areas very carefully.

In a second experiment, we draw two boxplots to show the
percent of normal objects (i.e., the success rate) within the
defensive walls as shown in
Figure~\ref{fig:boxplot}. We again have 100 runs. For  
each run, we randomly select 100 points to keep their labels. Based on
the labels, we perform ADClust to 
cluster instances. Then we
set different weights and examine different alpha levels
for the defensive walls. We set $k$ to 1, 30 and 50 as low, medium and
high weights. For each of the weight, we show the
success rates for the two types of defensive walls. 

We set $\alpha$ levels from 0.6 to
0.95. The median of the success rates varies from 0.85
to 0.87. We find that the weights $k=30$ and
$k=50$ perform better than $k=1$ in term of success 
rate. Furthermore, $\alpha =0.8$ has the highest median success
rate for Manhattan defensive walls and $\alpha = 0.7$ has the
highest median success rate for Euclidean defensive walls. 
Both of the results are consistent with the recommended 
$\alpha$ range, 0.6 to 0.8, from the game theoretic studies.

Note semi-supervised learning techniques are designed to
achieve the highest overall accuracy over all unlabeled normal and
abnormal objects. On the other hand, our algorithm does keep many points
unlabeled, hence we do not have an
overall accuracy measure computed over all the unlabeled 
objects. Meanwhile one of our algorithm's focus is to have
objects as purely normal inside the defensive walls as possible,
at the expense of decreased accuracy, since many normal
objects are blocked out of the wall along with the abnormal
ones. In this experiment, KDD data is a highly mixed data,
yet we achieve on average nearly 90\% pure normal rate inside the
defensive walls, marked as the relatively safe regions. 
The unlabeled mixed regions, and
unlabeled whole clusters if there is any, are 
another focus of our algorithm.   
Results from our algorithm can be used for
tiered screenings of the objects, with the objects in the mixed
region examined most carefully to separate normal from abnormal,
and the unknown clusters examined for potential new attacks.   

%MK: Again from accuracy point of view, how does it compare to semi-supervised learning ??
%%
\section{Conclusion}
\label{sec:conclusion}
In this paper, we develop a novel adversarial clustering
algorithm, a.k.a. ADClust, to separate the attack region and the
normal region within mixed clusters caused by
adversaries' attack objects. With very few labeled
instances, we cannot build an effective classifier, which has a
clearly defined classification boundary to defend the normal
population from the attack objects. However utilizing the few
labeled objects, our clustering algorithm can identify
the mixed area between the normal region and the
attack region. Instead of a classifier boundary, analogous to a
point estimate, an
overlapping area is analogous to a confidence region, showing 
the strength of an attack.   
Furthermore defensive walls are
drawn inside normal regions. This is a
conservative strategy to defend the normal population against
active adversaries. All 
objects outside the centers of the normal objects need to
be examined carefully, especially in the mixed regions. \\ 

\noindent
{\bf Acknowledgements} This work is supported in part by ARO
grant W911NF-17-1-0356, NIH award 1R01HG006844 and NSF
CNS-1111529, CNS-1228198, CICI-1547324, and IIS-1633331.

\end{document}